\documentclass[letterpaper]{article}
\usepackage[preprint]{aaai2027}
\usepackage[hyphens]{url}
\usepackage{graphicx}
\usepackage{natbib}
\usepackage{caption}
\usepackage{algorithm}
\usepackage{algorithmic}
\usepackage{newfloat}
\usepackage{listings}
\DeclareCaptionStyle{ruled}{labelfont=normalfont,labelsep=colon,strut=off}
\floatstyle{ruled}
\newfloat{listing}{tb}{lst}{}
\floatname{listing}{Listing}
\usepackage{booktabs}
\usepackage{amssymb}
\usepackage{amsmath}
\usepackage{multirow}
\usepackage[table]{xcolor}
\title{TReVS: Integrating Textual Relevance and Visual Saliency for Efficient Vision-Language Model Token Pruning}
\author{
    Jing Wang\textsuperscript{\rm 1}\equalcontrib,
    Zhiping Wu\textsuperscript{\rm 2}\equalcontrib,
    Dongdong Ren\textsuperscript{\rm 3},
    Youfang Han\textsuperscript{\rm 4},
    Wei Zhao\textsuperscript{\rm 4},
    Wenbin Li\textsuperscript{\rm 1}\corresponding
}
\affiliations{
    \textsuperscript{\rm 1}School of Intelligence Science and Technology, Nanjing University\\
    \textsuperscript{\rm 2}School of Electronic Science and Engineering, Nanjing University\\
    \textsuperscript{\rm 3}Geely Automobile Research Institute (Ningbo) Co., Ltd., 315000\\
    \textsuperscript{\rm 4}Alpha Labs, Goertek\\
    {\footnotesize Jing Wang: 221900040@smail.nju.edu.cn; Zhiping Wu: zhipingwu@smail.nju.edu.cn;\\
    Dongdong Ren: Dongdong.Ren2@geely.com; Youfang Han: fred.hanyf@goertek.com;\\
    Wei Zhao: charles.zhaow@goertek.com; Wenbin Li: liwenbin@nju.edu.cn}
}

\begin{document}

\maketitle

\begin{abstract}

Vision-Language Models (VLMs) excel at visual understanding and reasoning but often incur substantial inference costs due to the large number of visual tokens. Recent visual token pruning methods increasingly follow a two-stage paradigm: they first remove visually redundant tokens after the vision encoder and then discard tokens irrelevant to the textual query within the Large Language Model (LLM). However, since the first stage typically relies solely on vision-encoder saliency, it may prematurely eliminate query-relevant tokens, depriving the subsequent text-guided stage of critical visual evidence. Our empirical analysis shows that incorporating query guidance into first-stage pruning better preserves task-relevant evidence and consistently improves performance over vision-only saliency-based pruning. We further find that high-variance attention heads are more sensitive to the textual query and yield more discriminative text-to-vision attention signals for second-stage pruning. Motivated by these findings, we propose \textbf{TReVS}, a training-free framework that combines textual relevance with vision-encoder saliency for pre-LLM pruning and leverages high-variance attention heads to remove task-irrelevant tokens at shallow-to-intermediate layers of the LLM. On LLaVA-1.5-7B, TReVS retains 92.8\% of the unpruned baseline performance while pruning 94.4\% of visual tokens, outperforming prior state-of-the-art methods.

\end{abstract}

\section{Introduction}
\label{intro}

\begin{figure}[!th]
\centering
\includegraphics[width=0.85\linewidth]{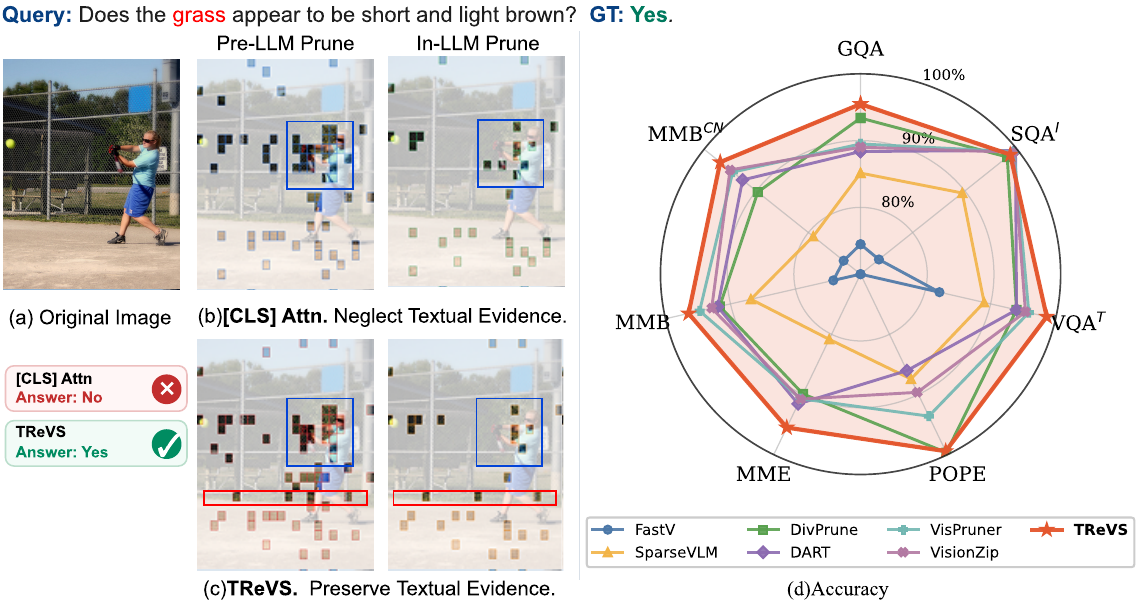}
\caption{
(a--c) A two-stage baseline guided only by [CLS] attention during pre-LLM pruning discards visual tokens representing the queried grass, leaving incomplete evidence for subsequent query-guided pruning. TReVS uses textual relevance in pre-LLM pruning, preserving this evidence for the in-LLM stage. (d) TReVS achieves the best performance across six image-understanding benchmarks.}

\label{fig:performance}
\end{figure}

Vision-Language Models (VLMs) extend the reasoning  of pretrained Large Language Models (LLMs)~\citep{touvron2023llama,bai2023qwen,achiam2023gpt} to visual inputs, achieving strong performance in visual question answering, multimodal reasoning, high-resolution image understanding, and video understanding~\citep{liu2023visual,zhu2023minigpt,li2023blip,li2024llava}. Recent VLMs improve fine-grained perception across high-resolution images, multiple images, and videos. This broader coverage, however, produces visual-token sequences substantially longer than text sequences. LLaVA-1.5 encodes each image into 576 visual tokens~\citep{liu2024improved}, LLaVA-NeXT uses up to 2,880 tokens per high-resolution image~\citep{liu2024llavanext}, and Qwen2.5-VL processes up to 16,384 visual tokens for multi-image and video inputs~\citep{Qwen2.5-VL}. These sequences increase computation, latency, and memory use, making visual-token processing a major bottleneck in efficient VLM inference.

Visual token pruning addresses this bottleneck by exploiting the substantial redundancy in visual inputs~\citep{bolya2023tome,chen2024image,zhang2024sparsevlm,yang2025visionzip}. Recent methods increasingly adopt a two-stage design that reduces visual redundancy after the vision encoder and then removes textually irrelevant tokens within the LLM~\citep{takezoe2026learnpruner,zhang2026vscan,singh2026duetvlm}. The pre-LLM stage typically uses vision-encoder saliency or diversity to select tokens. The in-LLM stage uses text-to-vision cross-attention to retain query-relevant tokens. Although this division improves the performance, it creates an information bottleneck. Current pre-LLM pruning remains largely query-agnostic and can discard visual evidence that is essential to the text query. Once removed, such evidence is unavailable to subsequent query-aware pruning. The second stage must therefore identify task-relevant tokens within a potentially incomplete visual context.

\begin{figure*}[!th]
\centering
\includegraphics[width=0.85\linewidth]{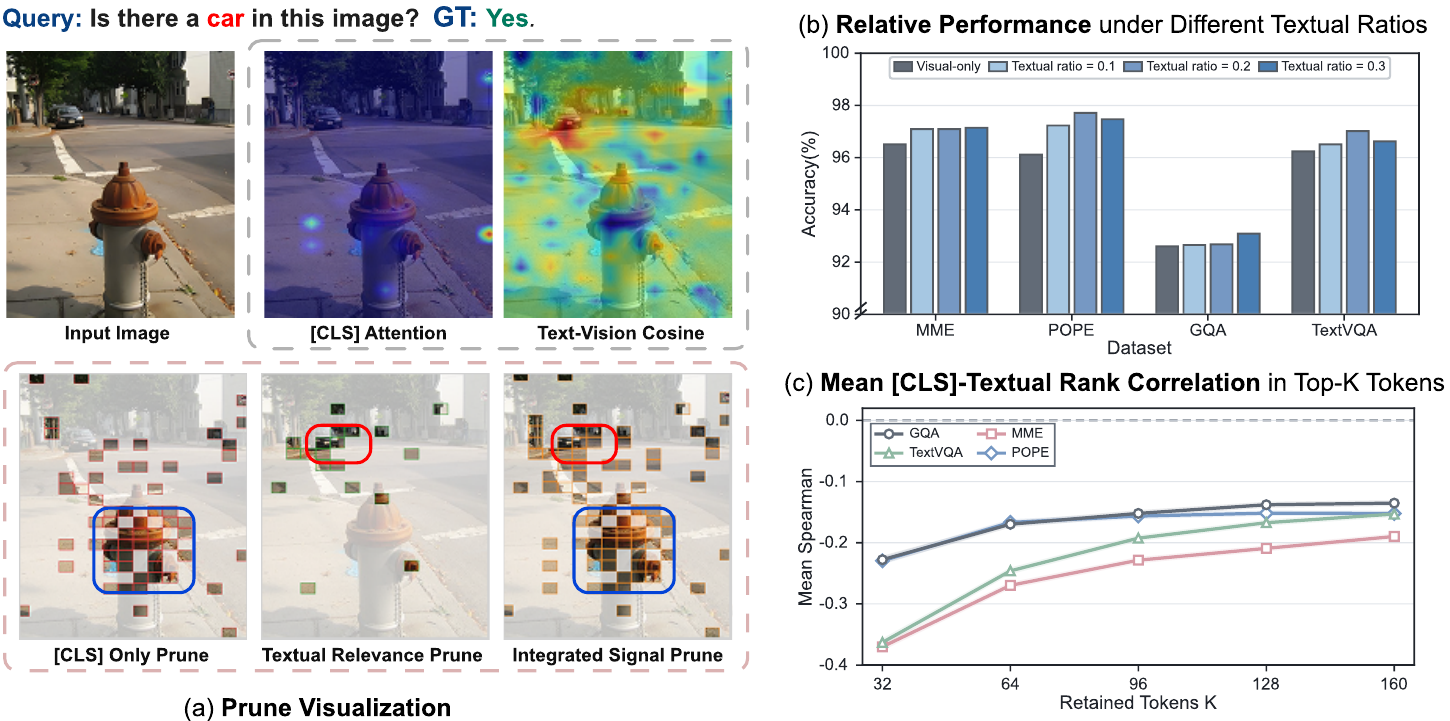}
\caption{Effect of textual guidance on pre-LLM pruning. (a) Vision-encoder saliency focuses on the dominant fire hydrant, whereas textual relevance identifies the queried car. Their integration preserves both. (b) Performance relative to dense inference under different textual ratios. (c) Mean Spearman correlation between the two signals within the top-$K$ saliency-selected tokens.}
\label{fig:rescue}
\end{figure*}

We first examine the information loss caused by query-agnostic pre-LLM pruning and obtain two findings. First, incorporating textual relevance into pre-LLM pruning consistently improves performance by preserving query-relevant evidence (Figure~\ref{fig:rescue}(b)). Second, vision-encoder saliency and textual relevance produce distinct token rankings, with negative Spearman correlations across datasets and token budgets (Figure~\ref{fig:rescue}(c)). Together, these results show that textual relevance complements vision-encoder saliency and should be introduced before irreversible token reduction. We then examine the in-LLM stage, where the retained visual tokens are further refined using text-to-vision attention. We find that high-variance attention heads produce more discriminative and query-sensitive attention signals for identifying task-relevant tokens. These findings motivate a coordinated two-stage design: the first stage preserves visually salient, textually relevant, and diverse evidence, while the second uses high-variance heads to remove residual task-irrelevant tokens after cross-modal interaction develops. Section~\ref{sec:preliminary analysis} provides detailed evidence for these findings.

As illustrated in Figure~\ref{fig:performance}, we propose \textbf{TReVS}, a training-free framework that coordinates pre-LLM and in-LLM visual token pruning. Before the LLM, TReVS scores visual tokens using [CLS] attention from the penultimate ViT layer and cosine similarity between projected visual tokens and embedded text tokens. It fuses these scores and supplements the top-ranked tokens with a small diversity set. During LLM prefilling, TReVS prunes again at a shallow-to-middle layer, using high-variance text-to-vision attention heads to identify query-relevant tokens. The first stage reduces visual redundancy while preserving salient, textually relevant, and diverse evidence. The second removes the remaining task-irrelevant tokens after cross-modal interaction develops.

Extensive experiments across vision-language benchmarks show that TReVS outperforms state-of-the-art training-free methods at multiple reduction ratios. On LLaVA-1.5-7B, TReVS prunes 94.4\% of visual tokens while retaining 92.8\% of the original performance. Without additional training, it accelerates prefilling by $2.2\times$ and end-to-end inference by $1.4\times$. These results establish the importance of preserving textual relevance before  visual token reduction.

Our main contributions are summarized as follows:
\begin{itemize}
    \item We show that introducing textual relevance before the first irreversible visual-token reduction preserves query-relevant evidence and consistently improves performance. The rankings induced by textual relevance and vision-encoder saliency remain negatively correlated across datasets and token budgets, showing that the two signals favor complementary visual tokens.
    
    \item We show that attention variance provides a simple, training-free criterion for identifying query-sensitive attention heads. These heads produce more discriminative text-to-vision attention signals for in-LLM pruning.
    
    \item We introduce TReVS, a training-free two-stage framework that combines vision-encoder saliency, textual relevance, and token diversity before the LLM, then uses high-variance attention heads to remove the remaining task-irrelevant visual tokens within the LLM. TReVS retains 92.8\% of the baseline performance while pruning 94.4\% of visual tokens.
\end{itemize}

\begin{figure*}[!th]
\centering
\includegraphics[width=0.85\linewidth]{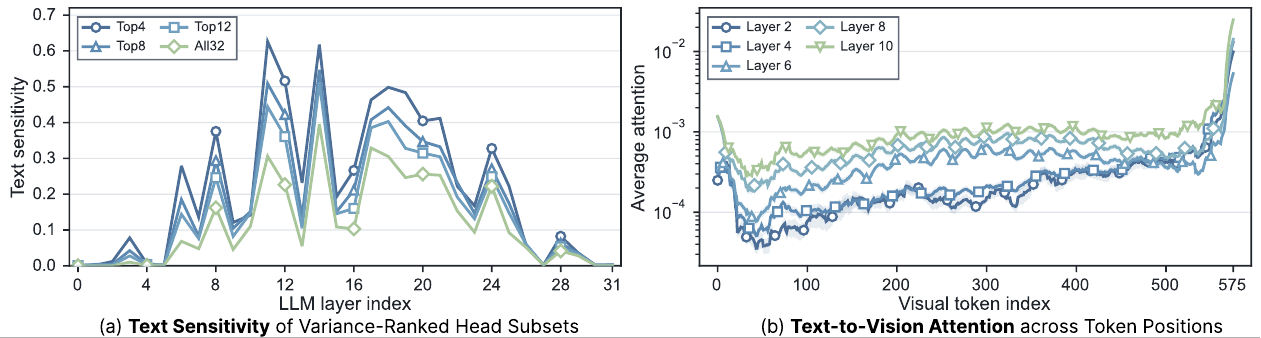}
\caption{Head sensitivity and positional attention patterns. (a) Mean textual sensitivity of cumulative head subsets ranked by text-to-vision attention variance. (b) Average text-to-vision attention across visual-token positions at selected LLM layers.}

\label{fig:priority}
\end{figure*}

\section{Related Work}
\label{sec:related_work}

\subsection{Vision-Language Models}

Vision-Language Models (VLMs) connect pretrained LLMs with vision encoders through projection modules such as MLPs and Q-Formers~\citep{liu2023visual,zhu2023minigpt,li2023blip}. Recent VLMs extend visual understanding from standard images to high-resolution images and long videos~\citep{liu2024improved,liu2024llavanext,Qwen2.5-VL,clark2026molmo2,liu2026kangaroo}. As image resolution and video duration increase, the resulting visual sequences can substantially exceed the corresponding text sequences, making visual token reduction important for efficient VLM inference.

\subsection{Visual Token Reduction for VLMs}

Existing methods differ primarily in where token reduction occurs and when the text query becomes available.

\textbf{Vision-guided pre-LLM reduction.}
These methods remove redundant visual tokens after the vision encoder using visual-side signals such as saliency, similarity, or diversity~\citep{alvar2025divprune}. VisionZip~\citep{yang2025visionzip} retains dominant tokens according to vision-encoder saliency and merges residual tokens into contextual representations. VisPruner~\citep{zhang2025vispruner} combines vision-encoder attention with similarity-based deduplication. Although these methods reduce computation before the LLM, their selection remains independent of the text query.

\textbf{Query-aware in-LLM pruning.}
These methods prune within the LLM using text-to-vision cross-attention to retain task-relevant visual information. FastV~\citep{chen2024image} performs attention-based pruning in shallow LLM layers, PyramidDrop~\citep{xing2024pyramiddrop} progressively reduces tokens as visual redundancy increases with depth, and SparseVLM~\citep{zhang2024sparsevlm} combines text-guided attention with rank-adaptive sparsity. Since all visual tokens first enter the LLM, pruning often occurs in very shallow layers to provide meaningful computational savings. At these layers, text-to-vision cross-attention is susceptible to attention shift and dispersion~\citep{zhang2025vispruner}, which can cause substantial accuracy loss.

\textbf{Two-stage pruning.}
Recent methods combine pre-LLM visual reduction with query-guided refinement inside the LLM~\citep{zhang2026vscan,singh2026duetvlm,takezoe2026learnpruner}. DUET-VLM~\citep{singh2026duetvlm} first merges redundant tokens through local clustering and then performs layer-wise pruning using cross-modal attention. LearnPruner~\citep{takezoe2026learnpruner} replaces vision-encoder attention with a learnable pre-LLM importance predictor and performs text-guided pruning in intermediate LLM layers. This design reduces early computation while allowing the second stage to use more mature cross-modal interactions.

Existing two-stage methods typically separate visual-only pre-LLM reduction from query-aware in-LLM pruning. In contrast, TReVS coordinates query guidance across both stages: textual relevance complements vision-encoder saliency before the LLM, while query-sensitive attention further refines the retained tokens within the LLM. This design reduces visual redundancy without depriving the second stage of query-relevant context.

\section{Preliminary Analysis}
\label{sec:preliminary analysis}

\subsection{Does textual guidance improve performance?}
\label{sec:preliminary textual}

Two-stage methods typically use vision-encoder saliency or diversity signals for pre-LLM pruning and query-aware attention for in-LLM pruning. This mismatch creates an irreversible bottleneck: query-relevant evidence removed before the LLM cannot be recovered later. We therefore ask: \textit{Can textual guidance during pre-LLM pruning preserve such evidence and improve performance?}

We evaluate LLaVA-1.5-7B on MME~\cite{fu2023mme}, POPE~\cite{li2023evaluating}, TextVQA~\cite{singh2019towards}, and GQA~\cite{hudson2019gqa}. Under a fixed token budget, we vary the allocation between vision-encoder saliency, measured by [CLS] attention in the penultimate ViT layer, and textual relevance, measured by cosine similarity between projected visual tokens and text embeddings.

Figure~\ref{fig:rescue}(a) illustrates their complementary behavior. Vision-encoder saliency favors the dominant fire hydrant, while textual relevance recovers the car targeted by the query. Their integration retains both sources of evidence.

As shown in Figure~\ref{fig:rescue}(b), allocating even a modest fraction of the budget to textual relevance consistently outperforms visual-only pruning. Figure~\ref{fig:rescue}(c) further shows negative Spearman correlations between the two scores among the top-$K$ saliency-selected tokens across all datasets and budgets. Thus, textual relevance reorders even visually salient candidates and complements, rather than replaces, vision-encoder saliency. Together, these results show that textual guidance contributes complementary selection information and improves performance under a fixed token budget.

\subsection{Are High-Variance Attention Heads More Query-Sensitive?}
\label{sec:preliminary highvar}
Several in-LLM pruning methods average text-to-vision attention across all heads~\citep{chen2024image,takezoe2026learnpruner}, although heads differ in how strongly they respond to the query. We therefore ask: \textit{Can query-sensitive heads be identified with a training-free criterion?}

A concentrated attention distribution has high variance across visual tokens, but concentration alone does not imply query sensitivity. We therefore test whether attention variance ranks heads whose visual-token attention changes more with the question. We pair 1,000 TextVQA~\citep{singh2019towards} images with two distinct questions and process them using LLaVA-1.5-7B. At each layer, we rank the 32 heads by the variance of their text-to-vision attention and evaluate cumulative top-4, top-8, top-12, and all-head subsets. Let $\mathbf{s}_h^{(1)}$ and $\mathbf{s}_h^{(2)}$ be the visual-token attention distributions of head $h$ for the two questions. We define its textual sensitivity as
\begin{equation}
    \operatorname{TS}_h=1-\cos\!\left(\mathbf{s}_h^{(1)},\mathbf{s}_h^{(2)}\right).
\end{equation}

Figure~\ref{fig:priority}(a) shows that the highest-variance subsets are generally more text-sensitive, with the clearest separation in the middle layers. Since the subsets are cumulative, the decrease from top-4 to all 32 heads shows that adding lower-ranked heads reduces the mean per-head sensitivity. Attention variance therefore provides a simple criterion for selecting query-sensitive heads.

Figure~\ref{fig:priority}(b) compares average text-to-vision attention across visual-token positions at different layers. In the earliest layers, attention generally increases with token position, favoring visual tokens closer to the subsequent text sequence. Although all layers exhibit peaks near the sequence boundaries, later layers show flatter profiles across the interior positions, suggesting weaker positional dependence. This observation motivates delaying attention-based pruning beyond the earliest layers.

Together, these analyses motivate two design choices: using high-variance heads for query-sensitive token scoring and pruning at a shallow-to-middle layer to avoid strong early-layer position dependence.

\begin{figure*}[!th]
\centering
\includegraphics[width=0.85\linewidth]{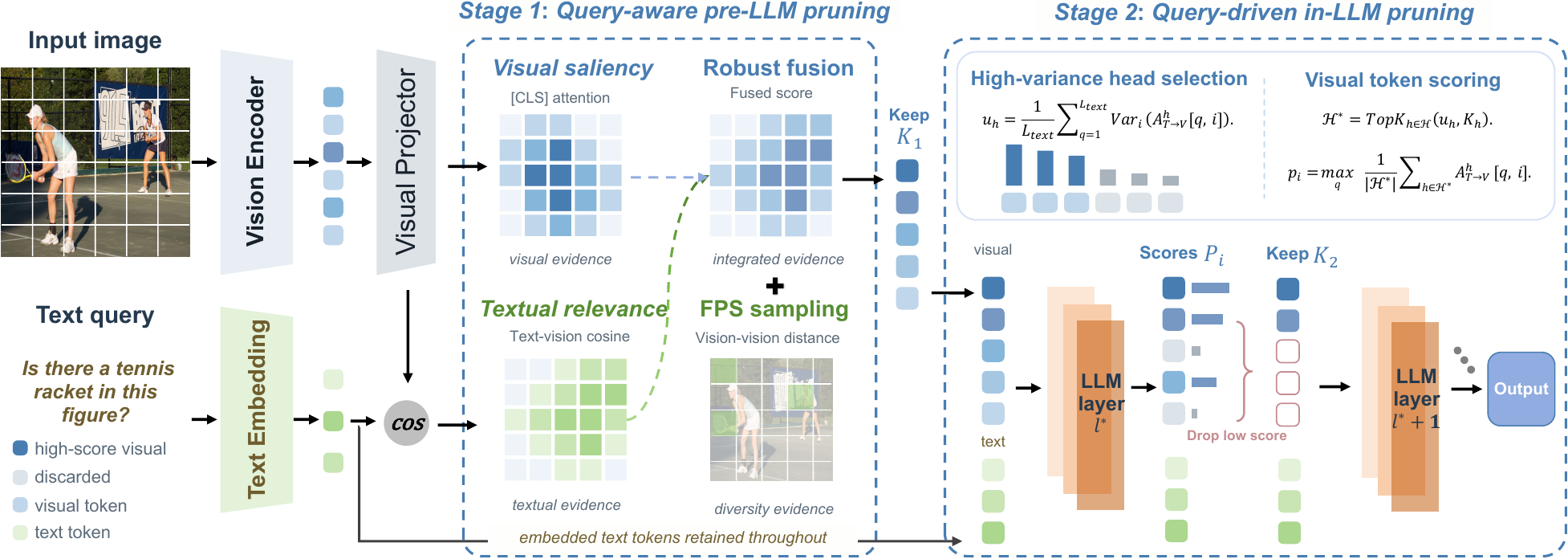}
\caption{The architecture of TReVS. Stage 1 performs query-aware pre-LLM pruning by preserving visually salient, query-relevant, and diverse tokens. Stage 2 uses high-variance attention heads to guide in-LLM visual token pruning.}

\label{fig:framework}
\end{figure*}

\section{Method}
\label{method}
In this section, we present \textbf{TReVS}, a training-free framework that coordinates pre-LLM and in-LLM visual token pruning (Figure~\ref{fig:framework}). Stage 1 retains $K_1$ visually salient, textually relevant, and diverse tokens before the LLM. Stage 2 uses high-variance attention heads to retain $K_2$ query-relevant tokens at a shallow-to-middle LLM layer. The following subsections detail the two stages.

\subsection{Query-Aware Visual Redundancy Reduction}

\textbf{Visual saliency score.} To preserve visually salient information, we follow prior works~\citep{yang2025visionzip} and leverage [CLS]-attention from the penultimate ViT layer as the vision-encoder saliency signal. Let $x_{[\mathrm{CLS}]} \in \mathbb{R}^{d}$ denote the [CLS] token, $X_v \in \mathbb{R}^{N \times d}$ the $N$ patch token features output by the ViT, and $W_Q^h, W_K^h \in \mathbb{R}^{d \times d_h}$ the query and key projection matrices for head $h$, where $d_h = d / H$ is the per-head dimension. The [CLS]-to-patch attention for head $h$ is:

\begin{equation}
\begin{aligned}
    q_{[\mathrm{CLS}]} = x_{[\mathrm{CLS}]} W_Q^h,
    \quad K_v = X_v W_K^h, \\
    A_{[\mathrm{CLS}]}^h = \mathrm{Softmax}\!\left(\frac{q_{[\mathrm{CLS}]} K_v^\top}{\sqrt{d_h}}\right).
\end{aligned}
\end{equation}

The visual saliency score for the $i$-th patch token is obtained by averaging the [CLS]-attention across all $H$ heads:

\begin{equation}
    s_i^v = \frac{1}{H} \sum_{h=1}^{H} A_{[\mathrm{CLS}]}^h[i].
\end{equation}

\textbf{Textual-relevance score.} As demonstrated in Section~\ref{sec:preliminary textual}, incorporating a text-guidance signal at the pre-LLM stage is critical for retaining the text-relevant visual tokens that the LLM stage requires. We first project the ViT output through the visual projector to obtain text-aligned representations: $Z = \mathrm{Projector}(X_v) \in \mathbb{R}^{N \times d'}$, where $z_i \in \mathbb{R}^{d'}$ is the projected feature of visual token $i$. Let $t_j \in \mathbb{R}^{d'}$ denote the embedding of the $j$-th text token, with $M$ tokens in total. We normalize the visual and text features as $\bar{z}_i = z_i / \lVert z_i \rVert_2$ and $\bar{t}_j = t_j / \lVert t_j \rVert_2$, respectively. We then compute the rectified cosine similarity between each text--visual pair:
\begin{equation}
    m_{j,i} = \max\!\left(\bar{t}_j^\top \bar{z}_i,0\right).
\end{equation}

The query-relevance score for visual token $i$ is the Root Mean Square (RMS) aggregation over all text tokens:
\begin{equation}
    s_i^t = \sqrt{\frac{1}{M}\displaystyle\sum_{j=1}^{M} m_{j,i}^2}.
\end{equation}

This aggregation suppresses background noise from weakly related text tokens while amplifying tokens that are broadly attended to by the query.

\textbf{Normalization and temperature scaling.} Since $s_i^v$ and $s_i^t$ are drawn from distributions with different scales and statistics, we independently apply robust normalization to each before fusion. We use the Median Absolute Deviation (MAD) as a spread estimator, which is resistant to the attention outliers commonly observed in [CLS] Attention:
\begin{equation}
    \hat{s}_i^c = \left[\frac{s_i^c - \mathrm{Med}(\mathbf{s}^c)}{\mathrm{MAD}(\mathbf{s}^c) + \epsilon}\right]_{\!+} \bigg/ \, \tau_c, \quad c \in \{v,\, t\}.
\end{equation}
Here, $[x]_+=\max(x,0)$ clamps negative values to zero. $\mathrm{Med}(\mathbf{s}^c)=\underset{i}{\operatorname{median}}(s_i^c)$ denotes the median score across all visual tokens, while $\mathrm{MAD}(\mathbf{s}^c)=\underset{i}{\operatorname{median}}\left(\left|s_i^c-\mathrm{Med}(\mathbf{s}^c)\right|\right)$ measures the median absolute deviation from this center. The constant $\epsilon$ ensures numerical stability, and $\tau_c$ controls the temperature scaling, with $\tau_v=1.4$ and $\tau_t=1.0$ by default.

\textbf{Pivot token selection.} Visually salient tokens do not necessarily exhibit high textual relevance, suggesting that the two signals capture complementary evidence. We therefore define a unified score that preserves tokens favored by either signal and adds a reward when both scores are high:
\begin{equation}
    s_i = \max\!\left(\hat{s}_i^v,\, \hat{s}_i^t\right) + \lambda\sqrt{\hat{s}_i^v \cdot \hat{s}_i^t},
\end{equation}
where $\lambda\sqrt{\hat{s}_i^v \hat{s}_i^t}$ is the consistency reward, with $\lambda = 1.0$ by default. We retain the top-$K_r$ tokens as pivot tokens $\mathcal{S}_r = \operatorname{TopK}(s_i,\ K_r)$.

\textbf{Context preservation.} To mitigate the loss of background information, we sample $K_d$ tokens from the remaining candidates $\mathcal{C}=\{1,2,\ldots,N\}\setminus\mathcal{S}_r$ using Farthest Point Sampling (FPS). FPS iteratively selects the token farthest from the selected tokens in cosine distance, maximizing feature diversity. The resulting set is $\mathcal{S}_d=\operatorname{FPS}(\{z_i\mid i\in\mathcal{C}\},K_d)$. The union $\mathcal{S}_1 = \mathcal{S}_r \cup \mathcal{S}_d$ of $K_1 = K_r + K_d$ tokens is passed into the LLM for subsequent processing.

\subsection{Query-Driven Visual Token Compression}

After Stage 1 pruning, $K_1$ visual tokens are concatenated with the text tokens and fed into the LLM. As the forward pass progresses, the text tokens gradually absorb task-relevant visual information through cross-modal attention, causing the visual tokens to become increasingly redundant~\citep{kaduri2025whats}. To further reduce computational overhead, we perform a second pruning step at layer $l^*$, retaining $K_2$ visual tokens for subsequent layers.

\textbf{High-variance head selection.} As shown in Section~\ref{sec:preliminary highvar}, heads with higher text-to-vision attention variance are  more sensitive to changes in the text query. We therefore use attention variance as a training-free criterion for identifying query-sensitive heads at the second-stage pruning layer $l^*$. Let $A_{T \to V}^h \in \mathbb{R}^{L_\text{text} \times K_1}$ denote the text-to-vision attention matrix of head $h$ at this layer, where $q$ indexes text tokens and $i$ indexes visual tokens. The attention variance of head $h$ is:
\begin{equation}
u_h =
\frac{1}{L_\text{text}}
\sum_{q=1}^{L_\text{text}}
\operatorname{Var}_{i}\!\left(A_{T \to V}^h[q,\,i]\right).
\end{equation}
Here, $\operatorname{Var}_{i}$ computes the variance across the $K_1$ visual-token positions for each text token. Averaging over all text tokens yields $u_h$, which measures the overall visual-token selectivity of head $h$. We select the top half of the heads ranked by $u_h$ to retain query-sensitive attention signals while maintaining sufficient head coverage for stable visual-token scoring. The selected heads form the high-variance head set $\mathcal{H}^*$ used in the subsequent token selection.

\textbf{Query-irrelevant token reduction.} Using $\mathcal{H}^*$, we score each visual token by its maximum text-to-vision attention across all text positions, averaged over the selected heads:
\begin{equation}
    p_i = \max_q \; \frac{1}{|\mathcal{H}^*|} \sum_{h \in \mathcal{H}^*} A_{T \to V}^h[q,\, i].
\end{equation}

We retain the top-$K_2$ tokens by this score $\mathcal{S}_2 = \operatorname{TopK}(p_i,\ K_2)$.
The resulting $K_2$ visual tokens are then forwarded through the remaining LLM layers.

Overall, the two stages form a progressive query-aware pruning process. By introducing textual relevance into pre-LLM pruning, TReVS preserves query-relevant visual evidence before irreversible reduction, ensuring that the subsequent in-LLM stage operates on an informative visual context. Within the LLM, high-variance heads provide more discriminative text-to-vision attention, enabling the in-LLM stage to better retain task-relevant visual tokens. 

\section{Experiments}
\label{sec:experiments}

We evaluate TReVS across image, high-resolution, and video understanding to examine whether its query-aware two-stage design remains effective across different visual-token lengths. We then analyze its inference efficiency.

\subsection{Experimental Setup}
\label{sec:experimental_setup}

\textbf{Models and benchmarks.}
We evaluate TReVS on LLaVA-1.5-7B~\citep{liu2024improved}, LLaVA-NeXT-7B~\citep{liu2024llavanext}, and Video-LLaVA-7B~\citep{lin2023video}, which process 576, 2,880 and 2,048 visual tokens, respectively. For LLaVA-1.5-7B, we use GQA~\citep{hudson2019gqa}, ScienceQA-IMG~\citep{lu2022learn}, TextVQA~\citep{singh2019towards}, POPE~\citep{li2023evaluating}, MME~\citep{fu2023mme}, and the English and Chinese splits of MMBench~\citep{liu2024mmbench}. We evaluate LLaVA-NeXT-7B on GQA, TextVQA, MME, and MMBench, and Video-LLaVA-7B on TGIF-QA~\citep{jang2017tgif}, MSVD-QA, and MSRVTT-QA~\citep{xu2017video}.

\textbf{Implementation details.}
TReVS is training-free and leaves all parameters of the underlying VLM unchanged. Unless otherwise specified, we perform the second pruning stage after the eighth LLM layer and retain the top half of the attention heads according to their text-to-vision attention variance. We set the ratio between pivot and diversity tokens to $K_r:K_d=3:1$ and maintain $K_1:K_2=3:1$ between the two pruning stages. The visual and textual temperatures are set to $\tau_v=1.4$ and $\tau_t=1.0$, respectively. The consistency-reward weight is set to $\lambda=1.0$.

\begin{table*}[th]
\centering

\newcommand{\venueinfo}[1]{\textcolor{black!70}{\footnotesize~(#1)}}

\begin{tabular}{l|ccccccc|c}
\toprule
\textbf{Method} & \textbf{GQA} & \textbf{SQA$^{I}$} & \textbf{VQA$^{T}$} & \textbf{POPE} & \textbf{MME} & \textbf{MMB} & \textbf{MMB$^{CN}$} & \textbf{RelAcc.} \\
\midrule

\rowcolor{gray!25}\multicolumn{9}{c}{\textit{Upper Bound, 576 Tokens (100\%)}} \\
Vanilla & 61.9 & 69.5 & 58.2 & 85.9 & 1862 & 64.7 & 58.3 & 100.0\% \\

\rowcolor{gray!25}\multicolumn{9}{c}{\textit{Retain Averaged 128 Tokens (\textcolor{blue}{$\downarrow$77.8\%})}} \\
FastV \venueinfo{ECCV 2024} & 49.6 & 60.2 & 50.6 & 59.6 & 1490 & 56.1 & 51.4 & 82.6\% \\
SparseVLM \venueinfo{ICML 2025} & 56.0 & 67.1 & 54.9 & 80.5 & 1696 & 60.0 & 51.1 & 92.4\% \\
DivPrune \venueinfo{CVPR 2025} & 59.3 & \underline{69.0} & 56.1 & \textbf{86.7} & 1718 & 62.0 & 54.8 & 96.4\% \\
VisionZip \venueinfo{CVPR 2025} & 57.6 & 68.9 & 56.8 & 83.2 & 1762 & 62.0 & 56.7 & 96.3\% \\
VScan \venueinfo{TMLR 2026} & \underline{59.8} & 68.9 & \underline{57.3} & 86.1 & \underline{1792} & 63.0 & \textbf{58.0} & \underline{98.2\%} \\
DUET-VLM \venueinfo{CVPR 2026} & 59.0 & \textbf{70.2} & \textbf{57.8} & 85.9 & 1767 & \underline{63.3} & 56.7 & 97.9\% \\
TReVS & \textbf{60.3} & 68.9 & \textbf{57.8} & \underline{86.5} & \textbf{1842} & \textbf{63.6} & \underline{57.2} & \textbf{98.8\%} \\

\rowcolor{gray!25}\multicolumn{9}{c}{\textit{Retain Averaged 64 Tokens (\textcolor{blue}{$\downarrow$88.9\%})}} \\
FastV \venueinfo{ECCV 2024} & 46.1 & 51.1 & 47.8 & 48.0 & 1256 & 48.0 & 42.7 & 71.6\% \\
SparseVLM \venueinfo{ICML 2025} & 52.7 & 62.2 & 51.8 & 75.1 & 1505 & 56.2 & 46.1 & 85.4\% \\
DivPrune \venueinfo{CVPR 2025} & \underline{57.8} & 68.2 & 54.7 & \textbf{85.6} & 1674 & 59.3 & 52.3 & 93.8\% \\
VisionZip \venueinfo{CVPR 2025} & 55.1 & \underline{69.0} & 55.5 & 77.0 & 1690 & 60.1 & 55.4 & 93.1\% \\
VScan \venueinfo{TMLR 2026} & \textbf{58.3} & \textbf{69.1} & 55.6 & 85.0 & 1698 & \textbf{62.1} & 55.7 & \underline{95.8\%} \\
DUET-VLM \venueinfo{CVPR 2026} & 56.7 & 68.3 & \underline{56.4} & 82.5 & \underline{1751} & \underline{62.0} & \textbf{56.3} & 95.6\% \\
TReVS & \underline{57.8} & 68.9 & \textbf{56.6} & \underline{85.2} & \textbf{1769} & 61.9 & \underline{56.0} & \textbf{96.5\%} \\

\rowcolor{gray!25}\multicolumn{9}{c}{\textit{Retain Averaged 32 Tokens (\textcolor{blue}{$\downarrow$94.4\%})}} \\
FastV \venueinfo{ECCV 2024} & 41.5 & 42.6 & 42.5 & 32.5 & 1090 & 37.8 & 33.2 & 59.0\% \\
SparseVLM \venueinfo{ICML 2025} & 48.3 & 57.3 & 46.1 & 67.9 & 1290 & 51.4 & 40.6 & 76.7\% \\
DivPrune \venueinfo{CVPR 2025} & \underline{54.9} & 68.6 & 52.9 & \underline{81.5} & 1611 & 57.6 & 49.1 & 90.4\% \\
VisionZip \venueinfo{CVPR 2025} & 51.8 & 68.8 & 53.1 & 68.7 & 1536 & 57.7 & 50.3 & 87.4\% \\
VScan \venueinfo{TMLR 2026} & 54.8 & \textbf{69.4} & \underline{53.9} & 79.9 & 1598 & 59.5 & \underline{51.9} & 91.5\% \\
DUET-VLM \venueinfo{CVPR 2026} & 53.6 & \underline{69.1} & \textbf{54.7} & 74.8 & \underline{1633} & \underline{60.0} & \textbf{54.8} & \underline{91.6\%} \\
TReVS & \textbf{55.0} & 69.0 & \underline{53.9} & \textbf{82.7} & \textbf{1651} & \textbf{61.4} & 51.7 & \textbf{92.8\%} \\
\bottomrule
\end{tabular}

\caption{Performance comparison on LLaVA-1.5-7B under matched average token budgets. RelAcc. averages benchmark-wise performance relative to the unpruned model. Bold and underlined entries denote the best and second-best results.}
\label{tab:results_llava15}
\end{table*}

\subsection{Main Results}
\label{sec:main_results}

Tables~\ref{tab:results_llava15}, \ref{tab:results_llavanext}, and~\ref{tab:video-llava} compare TReVS with existing visual token pruning methods across image and video VLMs.

\textbf{Results on LLaVA-1.5-7B.}
As shown in Table~\ref{tab:results_llava15}, TReVS achieves the highest RelAcc. at all three budgets, retaining 98.8\%, 96.5\%, and 92.8\% of the unpruned performance. Notably, TReVS consistently obtains the best MME scores across these budgets,  which indicates stable preservation of visual evidence required for diverse perception and cognition tasks. Moreover, its margin over the second-best RelAcc. increases from 0.6\% at 128 tokens to 1.2\% at 32 tokens. This widening margin is consistent with the two-stage design of TReVS. Before the LLM, textual relevance prioritizes query-relevant tokens, reducing critical information loss when only a small number of tokens can be retained. Within the LLM, query-sensitive heads provide more discriminative attention for selecting task-relevant tokens, making more effective use of the limited remaining token budget.

\begin{figure}[!ht]
\centering
\includegraphics[width=0.85\linewidth]{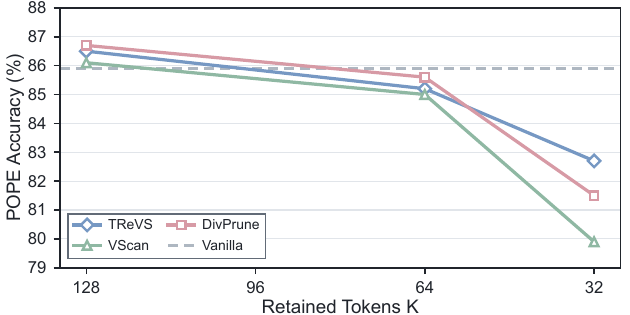}
\caption{POPE accuracy on LLaVA-1.5-7B across retained-token budgets. While all methods perform similarly at moderate budgets, TReVS degrades more slowly under increasingly aggressive compression and achieves the best performance at 32 tokens. The dashed line denotes the unpruned model.}

\label{fig:pope_robustness}
\end{figure}

\textbf{Hallucination robustness.}
Figure~\ref{fig:pope_robustness} compares POPE performance across token budgets. While the methods perform similarly at moderate budgets, TReVS achieves 82.7\% accuracy with 32 tokens, outperforming DivPrune and VScan by 1.2\% and 2.8\%. This robustness is consistent with our query-aware pre-LLM design, which preserves evidence for queried objects before irreversible pruning and thereby mitigates hallucinations caused by missing visual evidence.

\textbf{Results on LLaVA-NeXT-7B.}
Table~\ref{tab:results_llavanext} evaluates TReVS on the longer visual sequences produced by high-resolution inputs. TReVS achieves the highest RelAcc. at both budgets, retaining 96.6\% and 93.0\% of the unpruned performance with 320 and 160 tokens, respectively. The best MME scores at both budgets and the best MMB score at 160 tokens further show that TReVS preserves key visual evidence under substantial compression. Despite these overall gains, TReVS trails the best GQA results by 0.5\% and 0.6\% under high-resolution inputs, whereas the gap is less pronounced with lower-resolution inputs. GQA requires global reasoning over multiple objects and their relations, and such evidence may be distributed across longer visual sequences. We hypothesize that high-variance heads prioritize sparse query-relevant details and therefore retain slightly less global context, making this trade-off more evident at higher resolutions.

\begin{table}[!th]
\centering
{
\setlength{\tabcolsep}{4pt}
\begin{tabular}{l|cccc|c}
\toprule
\textbf{Method}
& \textbf{GQA}
& \textbf{VQA$^{T}$}
& \textbf{MME}
& \textbf{MMB}
& \textbf{RelAcc.} \\
\midrule

\rowcolor{gray!25}
\multicolumn{6}{c}{\textit{Upper Bound, 2,880 Tokens (100\%)}} \\
Vanilla
& 64.2 & 61.3 & 1842 & 67.9 & 100.0\% \\
\rowcolor{gray!25}
\multicolumn{6}{c}{\textit{Retain Averaged 320 Tokens
(\textcolor{blue}{$\downarrow$88.9\%})}} \\
SparseVLM
& 57.7 & 55.9 & 1694 & 64.3 & 91.9\% \\
DivPrune
& \underline{61.1} & 56.2 & 1724 & 63.9 & 93.6\% \\
VisionZip
& 59.3 & 58.9 & 1702 & 63.1 & 93.4\% \\
DUET-VLM
& 60.6 & \textbf{59.9} & \underline{1788}
& 64.3 & 96.0\% \\
VScan
& \textbf{61.4} & \underline{59.4} & 1775
& \textbf{65.5} & \underline{96.3\%} \\
TReVS
& 60.9 & 59.3 & \textbf{1826}
& \underline{64.9} & \textbf{96.6\%} \\

\rowcolor{gray!25}
\multicolumn{6}{c}{\textit{Retain Averaged 160 Tokens
(\textcolor{blue}{$\downarrow$94.4\%})}} \\
SparseVLM
& 51.2 & 46.4 & 1542
& \underline{63.1} & 83.0\% \\
DivPrune
& \underline{59.3} & 54.1 & 1643
& 62.9 & 90.6\% \\
VisionZip
& 55.5 & 56.2 & 1630
& 60.1 & 88.8\% \\
DUET-VLM
& 58.6 & \underline{57.2} & 1686
& 62.9 & 92.2\% \\
VScan
& \textbf{59.6} & \textbf{57.7} & \underline{1699}
& 62.0 & \underline{92.6\%} \\
TReVS
& 59.0 & 56.9 & \textbf{1717}
& \textbf{63.9} & \textbf{93.0\%} \\
\bottomrule
\end{tabular}}
\caption{Performance comparison on LLaVA-NeXT-7B under matched average token budgets. }
\label{tab:results_llavanext}
\end{table}

\textbf{Results on Video-LLaVA-7B.}
Table~\ref{tab:video-llava} evaluates whether TReVS generalizes from images to longer video inputs. With 93.4\% of visual tokens removed, TReVS achieves the best result on all three benchmarks and retains 99\% of the average unpruned performance, exceeding DUET-VLM by 3.2\%. These results demonstrate that query-aware two-stage pruning remains effective for highly redundant video sequences.

\begin{table}[th]
\centering
\setlength{\tabcolsep}{6.5pt}
{\small
\begin{tabular}{l|ccc|c}
\toprule
\textbf{Method} & \textbf{TGIF} & \textbf{MSVD} & \textbf{MSRVTT} & \textbf{RelAcc.} \\
\midrule

\rowcolor{gray!25}\multicolumn{5}{c}{\textit{Upper Bound, 2,048 Tokens (100\%)}} \\
Video-LLaVA & 48.7 & 70.1 & 57.4 & 100.0\% \\

\rowcolor{gray!25}\multicolumn{5}{c}{\textit{Retain Averaged 136 Tokens (\textcolor{blue}{$\downarrow$93.4\%})}} \\
FastV & 30.4 & 44.3 & 39.4 & 64.8\% \\
SparseVLM & 44.7 & \underline{68.2} & 31.0 & 81.0\% \\
VisionZip & 42.4 & 63.5 & 52.1 & 89.5\% \\
DUET-VLM & \underline{46.7} & 68.0 & \underline{54.2} & \underline{95.8\%} \\
TReVS & \textbf{48.9} & \textbf{69.2} & \textbf{56.2} & \textbf{99.0\%} \\
\bottomrule
\end{tabular}
}
\caption{Performance comparison on Video-LLaVA-7B with 136 average retained tokens.}
\label{tab:video-llava}
\end{table}

\subsection{Ablation Study}

Table~\ref{tab:trevs_reliability_ablation} isolates the contributions of the two pruning stages under the 32-token preset. Replacing [CLS]-attention-only pre-LLM pruning with the first stage of TReVS improves RelAcc. from 91.8\% to 92.7\% while retaining all attention heads in the second stage. This gain confirms that combining textual relevance with visual saliency and token diversity better preserves query-relevant and complementary evidence before irreversible token reduction. Selecting high-variance heads further increases RelAcc. to 93.2\%, demonstrating that high-variance heads provide more discriminative query-aware signals than uniformly using all heads. Together, the two stages improve RelAcc. by 1.4\%, confirming their complementary roles in preventing premature information loss and removing residual task-irrelevant tokens.

\begin{table}[!b]
    \centering
    {\small
    \begin{tabular}{llc}
        \toprule
        \textbf{Stage 1}
            & \textbf{Stage 2}
            & \textbf{RelAcc. (\%)} \\
        \midrule
        {[CLS]} Attn
            & All Heads
            & 91.8 \\
        \midrule
        \multirow{2}{*}{TReVS}
            & All Heads
            & \underline{92.7} \\
            & High-Variance Heads
            & \textbf{93.2} \\
        \bottomrule
    \end{tabular}}
    \caption{Ablation study of the two-stage token-pruning strategies
    on LLaVA-1.5-7B under the 32-token preset. RelAcc. denotes the
    average relative accuracy over TextVQA, MMBench, GQA, and POPE.}
    \label{tab:trevs_reliability_ablation}
\end{table}

\subsection{Efficiency Analysis}
\label{sec:efficiency}

\textbf{Computational efficiency.}
Table~\ref{tab:efficiency_pope_llava15_7b} reports the inference efficiency of TReVS on POPE with LLaVA-1.5-7B, measured on an NVIDIA \textbf{RTX 4090} GPU. At 32 tokens, TReVS achieves a $2.2\times$ prefill speedup and a $1.4\times$ end-to-end speedup while reducing KV-cache consumption by $6.5\times$. Together with the accuracy retained under the same budget in Table~\ref{tab:results_llava15}, these results demonstrate a favorable accuracy--efficiency trade-off under aggressive compression.

\begin{center}
{\small
\setlength{\tabcolsep}{3pt}
\begin{tabular}{l|c c c c}
\toprule
\textbf{Method} & \textbf{Token}  & \textbf{Total Time}  & \textbf{Prefill Time} & \textbf{KV Cache (MB)} \\
\midrule
Vanilla
& 576
& 126.5 {\small (1.0$\times$)}
& 72.7 {\small (1.0$\times$)}
& 321.1 {\small (1.0$\times$)}\\
\midrule
\multirow{3}{*}{Ours}
& 128
& 100.7 {\small (1.3$\times$)}
& 40.6 {\small (1.8$\times$)}
& 99.7 {\small (3.2$\times$)}\\
& 64
& 90.4 {\small (1.4$\times$)}
& 33.7 {\small (2.2$\times$)}
& 66.2 {\small (4.9$\times$)}\\
& 32
& 88.9 {\small (1.4$\times$)}
& 33.5 {\small (2.2$\times$)}
& 49.7 {\small (6.5$\times$)}\\
\bottomrule
\end{tabular}}
\captionof{table}{Inference efficiency on POPE with LLaVA-1.5-7B. Latency is measured per sample in milliseconds, and parenthesized values report speedup or KV-cache reduction over dense inference.}
\label{tab:efficiency_pope_llava15_7b}
\end{center}

\section{Conclusion}

This work identifies an irreversible information bottleneck in two-stage visual token pruning: query-agnostic pre-LLM reduction can discard task-relevant evidence before query-aware reasoning begins. We introduce TReVS, a training-free framework that coordinates vision-encoder saliency, textual relevance, token diversity, and high-variance attention heads across pre-LLM and in-LLM pruning. Our findings highlight two key principles for effective visual token pruning: preserving query-relevant evidence before the LLM and using query-sensitive attention heads to identify task-relevant tokens within the LLM. Future work may explore adaptive token allocation between the pre-LLM and in-LLM stages based on input complexity and query demands.

\bibliography{aaai2027}

\end{document}